\def\year{2020}\relax
\documentclass[letterpaper]{article} 
\usepackage{aaai20}  
\usepackage{times}  
\usepackage{helvet} 
\usepackage{courier}  
\usepackage[hyphens]{url}  
\usepackage{graphicx} 
\usepackage{amsmath}
\usepackage{amsfonts}
\usepackage{booktabs}
\usepackage{microtype}
\usepackage{multirow}
\usepackage{nohyperref}
\title{Knowledge Graph Alignment Network with Gated Multi-hop \\Neighborhood Aggregation}
\author{
		Zequn Sun\textsuperscript{\rm 1}\footnotemark[1],
		Chengming Wang\textsuperscript{\rm 1}\thanks{This work was done while Zequn Sun was a research intern at Alibaba Group. The first two authors contributed equally.},
		Wei Hu\textsuperscript{\rm 1}\thanks{Corresponding author},
		Muhao Chen\textsuperscript{\rm 2},
		Jian Dai\textsuperscript{\rm 3},
		Wei Zhang\textsuperscript{\rm 3},
		Yuzhong Qu\textsuperscript{\rm 1} \\
	\textsuperscript{\rm 1}State Key Laboratory for Novel Software Technology, Nanjing University, China\\
	\textsuperscript{\rm 2}Department of Computer Science, University of California, Los Angeles, USA \\
	\textsuperscript{\rm 3}Alibaba Group, China \\
	\{zqsun,cmwang\}.nju@gmail.com, \{whu, yzqu\}@nju.edu.cn, muhaochen@ucla.edu, \{yiding.dj, lantu.zw\}@alibaba-inc.com
}
\begin{document}

\maketitle

\begin{abstract}
Graph neural networks (GNNs) have emerged as a powerful paradigm for embedding-based entity alignment due to their capability of identifying isomorphic subgraphs. However, in real knowledge graphs (KGs), the counterpart entities usually have non-isomorphic neighborhood structures, which easily causes GNNs to yield different representations for them. To tackle this problem, we propose a new KG alignment network, namely AliNet, aiming at mitigating the non-isomorphism of neighborhood structures in an end-to-end manner. As the direct neighbors of counterpart entities are usually dissimilar due to the schema heterogeneity, AliNet introduces distant neighbors to expand the overlap between their neighborhood structures. It employs an attention mechanism to highlight helpful distant neighbors and reduce noises. Then, it controls the aggregation of both direct and distant neighborhood information using a gating mechanism. We further propose a relation loss to refine entity representations. We perform thorough experiments with detailed ablation studies and analyses on five entity alignment datasets, demonstrating the effectiveness of AliNet.
\end{abstract}

\section{Introduction}
Entity alignment is the task of finding entities from different knowledge graphs (KGs) that refer to the same real-world identity. Recently, increasing attention has been paid to the utilization of KG representation learning rather than symbolic formalism for tackling this task. Representation learning models encode KGs into vector spaces, where relation semantics of entities can be assessed by the learned embedding operations, such as the relation-specific translation \cite{TransE} or rotation \cite{RotatE}. For embedding-based entity alignment, the similarity of entities is measured by the distance of entity embeddings. It has shown great potentials in dealing with the symbolic heterogeneity problem and benefits the entity alignment task in both~monolingual and cross-lingual scenarios~\cite{IPTransE,MTransE}.

Most recently, graph neural networks (GNNs) \cite{GCN,GAT,MixHop} have emerged as a powerful model to learn vector representations for graph-structured data. In GNNs, the representation of a node is learned by recursively aggregating the representations of its neighboring nodes. A recent work \cite{K-GNN} has proved that GNNs have the same expressiveness as the Weisfeiler-Leman (WL) test \cite{WL} in terms of identifying isomorphic subgraphs. It provides the theory basis of using GNNs for entity alignment between different KGs as similar entities usually have similar neighborhood. Recently, several studies \cite{GCN_Align,KGMatching,MuGNN,RDGCN,AVR-GCN} have exploited GNNs for embedding-based entity alignment, and have achieved promising results. 

\begin{figure}[!t]
	\center
	\includegraphics[width=0.9999\linewidth]{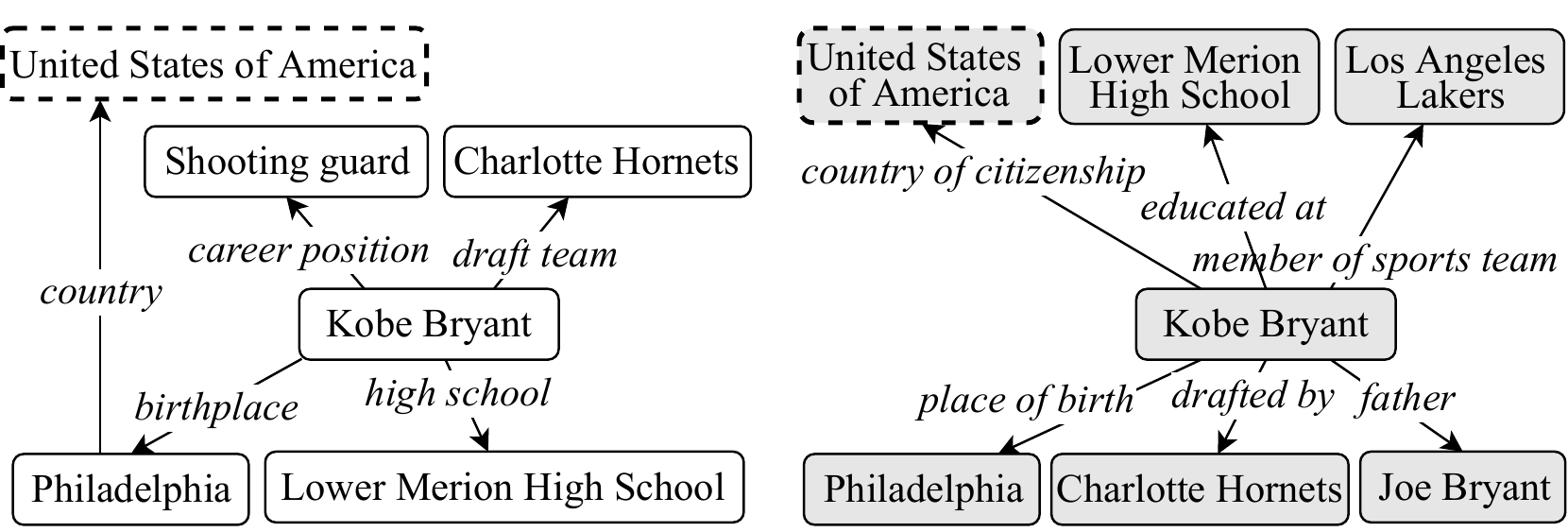}
	\caption{Non-isomorphic relational neighborhood of \textit{Kobe Bryant} in DBpedia (left) and Wikidata (right), respectively.}
	\label{fig:example}
\end{figure}

However, existing GNN-based entity alignment models still face a critical problem. As different KGs usually have heterogeneous schemas and data incompleteness \cite{KGIdentification}, the counterpart entities usually have dissimilar neighborhood structures. Figure~\ref{fig:example} gives an example. The neighborhood of the two entities referring to \textit{Kobe Bryant} is inconsistent to each other, especially containing different sets of neighboring entities. The statistics on DBpedia-based benchmark datasets for entity alignment~\cite{JAPE} also show that the majority of aligned entity pairs have different neighboring entities. Particularly, the percentages of such entity pairs reach $89.97\%$ between Chinese-English, $86.19\%$ between Japanese-English and $90.71\%$ between French-English, respectively. Different neighborhood structures would easily cause a GNN to yield different representations for counterpart entities. 

The challenge of resolving this issue lies in the difficulty of fully mitigating the non-isomorphism in the neighborhood structures of counterpart entities from different KGs. Even though we assume that the two KGs are complete (the goal of MuGNN \cite{MuGNN}), due to the schema heterogeneity, the counterpart entities still inevitably have dissimilar neighborhood structures. For example, in Figure~\ref{fig:example}, \textit{United States of America} is among the one-hop (direct) neighbors of \textit{Kobe Bryant} in Wikidata. However in DBpedia, it is a two-hop neighbor. Motivated by the fact that the semantically-related information can appear in both direct and distant neighbors of counterpart entities, we propose the KG alignment network AliNet which aggregates both direct and distant neighborhood information. Specifically, each AliNet layer has multiple functions to aggregate the neighborhood information within multiple hops. To reduce noise information, we further employ an attention mechanism for the distant neighborhood aggregation to find out important neighbors in an end-to-end manner. Finally, we use the gating mechanism to combine the output representations of the multiple aggregation functions, obtaining the hidden representations in the current layer. We also design a relation loss to refine entity representations and enable AliNet to capture some special structures such as the triangular relational structure. We perform thorough experiments with detailed ablation studies and analyses on five entity alignment datasets, demonstrating the effectiveness of AliNet and each of its technical contributions.

\begin{figure*}[!t]
	\center
	\includegraphics[width=0.95\linewidth]{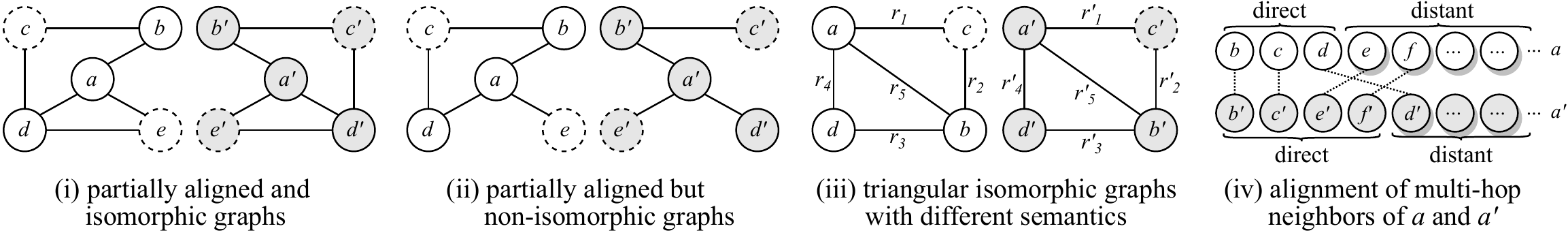}
	\caption{Illustration of GNNs for entity alignment. In (\romannumeral1), (\romannumeral2) and (\romannumeral3), $(a, a')$, $(b, b')$ and $(d, d')$ denote three pairs of pre-aligned entities while others are entities to be aligned. The dotted lines in (\romannumeral4) means the alignment relationship.}
	\label{fig:graph}
\end{figure*}

\section{Preliminaries}
\label{sect:preli}
\subsection{GNNs}
In GNNs, the representation of a node is learned by recursively aggregating the feature vectors of its neighbors. Different aggregation strategies lead to different variants of GNNs. 
\subsubsection{GCN} A very popular variant of GNNs is the vanilla GCN \cite{GCN}. The hidden representation of node $i$ at the $l$-th layer ($l\geq 1$), denoted as $\mathbf{h}_i^{(l)}$, is computed by: 
\begin{equation} 
	\label{eq:gcn}
	\mathbf{h}_i^{(l)} = \sigma \Big (\sum_{j\in \mathcal{N}_1(i)\cup \{i\} } \frac{1}{c_i} \mathbf{W}^{(l)} \mathbf{h}_j^{(l-1)} \Big ),
\end{equation}
where $\mathcal{N}_1(\cdot)$ represents the set of one-hop neighbors of the given entity, $\mathbf{W}^{(l)}$ is the weight matrix of the $l$-th layer and $c_i$ is the normalization constant. $\sigma(\cdot)$ is an activation function. The vanilla GCN encodes a node as the \textit{mean pooling} of the representations of its neighbors and itself from the last~layer. The input vector fed to the first layer is denoted as $\mathbf{h}_i^{(0)}$.

\subsubsection{R-GCN} Conventional GNNs only consider the node-wise connectivity in a graph and ignore edge labels such as the relations in KGs. R-GCN \cite{R-GCN} addresses this issue by distinguishing different neighbors with relation-specific weight matrices. It computes $\mathbf{h}_i^{(l)}$ as follows: 
\begin{equation} 
\label{eq:rrgcn}
\mathbf{h}_i^{(l)} = \sigma \Big (\mathbf{W}_0^{(l)}\mathbf{h}_i^{(l-1)} + \sum_{r\in \mathcal{R}}\sum_{j\in \mathcal{N}_r(i)} \frac{1}{c_{i,r}} \mathbf{W}_r^{(l)} \mathbf{h}_j^{(l-1)} \Big ),
\end{equation}
where $\mathbf{W}_0^{(l)}$ is the weight matrix for the node itself and $\mathbf{W}_r^{(l)} $ is used specifically for the neighbors having relation $r$, i.e., $\mathcal{N}_r(i)$. $\mathcal{R}$ is the relation set and $c_{i,r}$ is for normalization. 

\subsection{Entity Alignment of KGs}
We formally represent a KG as $\mathcal{G}=(\mathcal{E},\mathcal{R},\mathcal{T})$, where $\mathcal{E}$ is the set of entities, $\mathcal{R}$ is the set of relations, and $\mathcal{T}=\mathcal{E}\times\mathcal{R}\\\times\mathcal{E}$ is the set of triples. Without loss of generality, we consider the entity alignment task between two KGs, i.e., $\mathcal{G}_1=\\(\mathcal{E}_1,\mathcal{R}_1,\mathcal{T}_1)$ and $\mathcal{G}_2=(\mathcal{E}_2,\mathcal{R}_2,\mathcal{T}_2)$. Given partial pre-aligned entity pairs $\mathcal{A}^{+}=\{(i,j)\in\mathcal{E}_1 \times \mathcal{E}_2|i \equiv j\}$ where $\equiv$ means the alignment relationship, the goal of the task is to find alignment of remaining entities via entity embeddings. 

\subsection{GNNs for Entity Alignment}
Recent GNN-based entity alignment models include GCN-Align \cite{GCN_Align}, GMNN \cite{KGMatching}, MuGNN \cite{MuGNN}, RDGCN \cite{RDGCN} and AVR-GCN \cite{AVR-GCN}. GCN-Align and GMNN are built based on the vanilla GCN. RDGCN introduces dual relation graphs to enhance the vanilla GCN. AVR-GCN extends R-GCN using a TransE-like relation-specific translation operation \cite{TransE}. Before aggregation, each entity representation is translated from its tail entity representations using relation vectors. We argue that such relation-specific translation and R-GCN introduce a high complexity with the overhead of trainable parameters. More importantly, the aforementioned models do not take the non-isomorphism in KG structures into consideration. While MuGNN \cite{MuGNN} notices the structure incompleteness of KGs and proposes a two-step method of rule-based KG completion and multi-channel GNNs for entity alignment. However, the learned rules rely on relation alignment to resolve schema heterogeneity.

\subsubsection{Isomorphic structures are beneficial}
GNNs would learn the same representation for the entities that have isomorphic neighborhood structures with identical feature vectors representing corresponding neighbors \cite{PowerGCN}. We show that, in some cases, if two entities have isomorphic neighborhood structures and only partially pre-aligned neighbor representations, GNNs can also capture the similarity of other neighbors to be aligned. Figure \ref{fig:graph} (\romannumeral1) gives an example. For simplicity, here we consider a single-layer GCN. We can let pre-aligned entities have the same representation by minimizing their Euclidean distance, i.e., $\mathbf{h}_a^{(0)}=\mathbf{h}_{a'}^{(0)}$, $\mathbf{h}_b^{(0)}=\mathbf{h}_{b'}^{(0)}$ and $\mathbf{h}_d^{(0)}=\mathbf{h}_{d'}^{(0)}$ as well as $\mathbf{h}_a^{(1)}=\mathbf{h}_{a'}^{(1)}$, $\mathbf{h}_b^{(1)}=\mathbf{h}_{b'}^{(1)}$ and $\mathbf{h}_d^{(1)}=\mathbf{h}_{d'}^{(1)}$ in the ideal condition. By the mean-pooling based aggregation, we have $\mathbf{h}_b^{(1)}= \sigma(\mathbf{W}^{(1)}(\mathbf{h}_b^{(0)}+\mathbf{h}_a^{(0)}+\mathbf{h}_c^{(0)})/3) $ and $\mathbf{h}_{b'}^{(1)}= \sigma(\mathbf{W}^{(1)}(\mathbf{h}_{b'}^{(0)}+\mathbf{h}_{a'}^{(0)}+\mathbf{h}_{c'}^{(0)})/3) $, yielding $\mathbf{h}_c^{(0)}=\mathbf{h}_{c'}^{(0)}$. Finally, the counterpart entities would have the same representation. This indicates that the alignment information between entities can be propagated across the different GNN layers and different isomorphic graphs given partially pre-aligned neighborhood. However, for entity alignment between different KGs, it is impossible to require the two KGs to have isomorphic structures due to the schema heterogeneity. Figure \ref{fig:graph} (\romannumeral2) gives an example of non-isomorphic graph structures, where $c$ and $c'$ would have different representations due to their different neighborhood structures. 

\subsubsection{Only structures are not enough} Conventional GNNs fall short of characterizing  some special subgraph structures such as triangular graphs. Figure \ref{fig:graph} (\romannumeral3) shows a simple example. In this case, if we use mean-pooling aggregation, we would get that $\mathbf{h}_a^{(1)}=\mathbf{h}_{a'}^{(1)}=\mathbf{h}_b^{(1)}=\mathbf{h}_{b'}^{(1)}$ because the four entities have isomorphic neighborhood structures. While in fact, $a$ and $b$ are different entities with a specific relation. We should take relations into consideration. Although R-GCN \cite{R-GCN} considers relations in the aggregation function, it relies on relation alignment for identifying similar entities. Let us review Eq. (\ref{eq:rrgcn}). R-GCN needs to learn a weight matrix $\mathbf{W}_r$ for each relation $r$. If the relations of two KGs are not pre-aligned (e.g., $r_1 \equiv r'_1$ and $r_2 \equiv r'_2$), the relation-specific aggregation functions in R-GCN would fall short of propagating the alignment information of entities.

\begin{figure*}[!t]
	\center
	\includegraphics[width=0.88\linewidth]{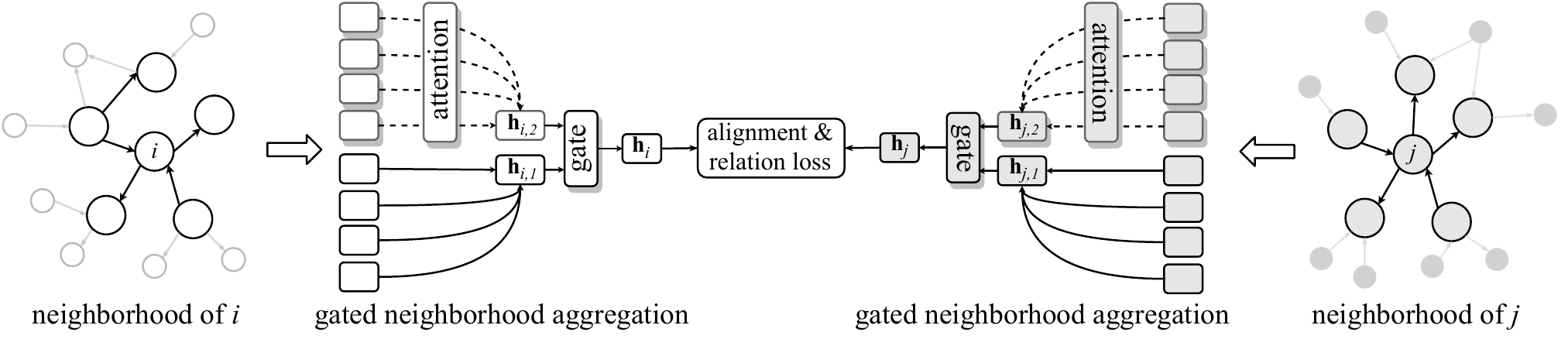}
	\caption{Overview of the KG alignment network (AliNet) with gated two-hop neighborhood aggregation.}
	\label{fig:alinet}
\end{figure*}

\subsubsection{Compensation with distant neighborhood and relations} The schema heterogeneity of different KGs usually brings about the mixture of direct and distant neighbors of counterpart entities. To reduce the effects of the non-isomorphism in neighborhood structures, we propose introducing distant neighborhood information. We show a toy example in Figure \ref{fig:graph} (\romannumeral4). The one-hop neighbors of two counterpart entities $a$ and $a'$ are different and only contain two pairs of counterpart entities ($b,b'$) and ($c,c'$). The one-hop neighbor $d$ of $a$ is in fact the distant neighbor $d'$ of $a'$. The distant neighbors $e$ and $f$ of $a$ are aligned with the one-hop neighbors $e'$ and $f'$ of $a'$, respectively. It is intuitive that if we can include the distant neighbors $e$ and $f$ in the neighborhood aggregation for $a$, and also take $d'$ into consideration for $a'$, the GNN would learn more similar representations for $a$ and $a'$. However, as can be seen, not all the distant neighbors are helpful. Therefore, the aggregation of distant neighbors should be attentive and selective. This is the key motivation of AliNet. To further enhance the expressiveness of AliNet, we also take relation semantics into consideration without introducing relation vectors.

\section{Knowledge Graph Alignment Network}

In AliNet, the entity representations are learned by a controlled aggregation of their neighborhood information within $k$ hops by the gating mechanism. Without loss of generality, in the following, we show the case of aggregating both the one-hop and two-hop neighborhood information ($k=2$). The network architecture is illustrated in Figure~\ref{fig:alinet}. Note that AliNet can also be extended to more hops.

\subsection{Gated Multi-hop Neighborhood Aggregation}
The one-hop neighbors of an entity are the most important neighborhood for GNNs to characterize the entity. We aggregate these neighbor representations using the vanilla GCN layers. Specifically, at the $i$-th layer, the hidden representation of entity $i$ by aggregating its one-hop neighbors, denoted as $\mathbf{h}_{i,1}^{(l)}$, can be computed using Eq. (\ref{eq:gcn}).

As discussed before, it is not enough to only aggregate one-hop neighbors. Although a GCN with $L$ layers can capture the structural information within the entity’s $L$-hop neighbors, such layer-by-layer propagation is not efficient. For two-hop neighborhood aggregation, we introduce the attention mechanism because directly employing the original aggregation of GCN would cause noise information to propagate through layers. Specifically, let $\mathcal{N}_2(\cdot)$ be the set of two-hop neighbors of the given entity. The hidden representation of entity $i$ by aggregating its two-hop neighborhood information at the $l$-th layer, denoted as $\mathbf{h}_{i,2}^{(l)}$, is computed as follows:
\begin{equation} 
\label{eq:alinet2}
\mathbf{h}_{i,2}^{(l)} = \sigma \Big (\sum_{j\in \mathcal{N}_2(i)\cup \{i\}} \alpha^{(l)}_{ij} \mathbf{W}_2^{(l)} \mathbf{h}_j^{(l-1)} \Big ), 
\end{equation}
where $\alpha^{(l)}_{ij}$ is a learnable attention weight for entity $i$ and its neighbor $j$. $\mathbf{W}_2^{(l)}$ is the weight matrix. The computation of attention weights is introduced in the next subsection.

Inspired by the skipping connections in neural networks \cite{HighwayNet,ResNet,RSN}. We propose to use the gating mechanism to combine the information from one-hop and two-hop neighbors directly. Specifically, the hidden representation $\mathbf{h}_{i}^{(l)}$of entity $i$ at the $l$-th layer is computed as follows:
\begin{equation} 
\label{eq:gating}
\mathbf{h}_{i}^{(l)} = g(\mathbf{h}_{i,2}^{(l)}) \cdot \mathbf{h}_{i,1}^{(l)} + (1-g(\mathbf{h}_{i,2}^{(l)})) \cdot \mathbf{h}_{i,2}^{(l)},
\end{equation}
where $g(\mathbf{h}_{i,2}^{(l)}) = \sigma(\mathbf{M} \mathbf{h}_{i,2}^{(l)} + \mathbf{b})$ serves as the gate to control the combination of both one-hop and two-hop neighborhood. $\mathbf{M}$ and $\mathbf{b}$ are the weight matrix and bias vector, respectively. 
\subsection{Attention for Distant Neighborhood}
The number of the more distant neighbors of an entity can grow exponentially than the number of its one-hop neighbors. It is intuitive that not all the distant neighbors contribute to the characterization of the central entity. Hence, for two-hop neighborhood aggregation, we compute the attention weights between entities for highlighting useful neighbors. The graph attention network GAT \cite{GAT} applies a shared linear transformation to entities in each attention function. However, as the central entity and its neighbors in KGs can be quite different, such shared transformation would cause a deleterious effect to correctly distinguishing between them. Instead, we use two matrices $\mathbf{M}_1^{(l)}$ and $\mathbf{M}_2^{(l)}$ for the linear transformations of the central entity and its neighbors, respectively. Formally, the attention weight $c_{ij}^{(l)}\in \mathbb{R}$ between $i$ and $j$ at the $l$-th layer is computed as follows:
\begin{equation} 
\label{eq:my_atten}
c_{ij}^{(l)} = \text{LeaklyReLU}[(\mathbf{M}_1^{(l)} \mathbf{h}_{i}^{(l)})^{\top}(\mathbf{M}_2^{(l)} \mathbf{h}_{j}^{(l)})],
\end{equation}
Finally, we normalize attention weights using the softmax function to make them comparable across different entities:
\begin{equation} 
\label{eq:atten_soft}
\alpha^{(l)}_{ij} = \text{softmax}_j(c^{(l)}_{ij})=\frac{\exp(c^{(l)}_{ij})}{\sum_{n\in\mathcal{N}_2(i)\cup \{i\}}\exp(c^{(l)}_{in})}.
\end{equation}

\subsection{Contrastive Alignment Loss}
We minimize the contrastive alignment loss to let the representations of aligned entities have a very small distance while those of unaligned entities have a large distance: 
\begin{equation}\label{eq:ali_loss}\small
	\mathcal{L}_1 = \sum_{(i,j) \in \mathcal{A}^{+}}||\mathbf{h}_i - \mathbf{h}_j|| + \sum_{(i',j') \in \mathcal{A}^{-}}\alpha_1[\lambda - || \mathbf{h}_{i'} - \mathbf{h}_{j'}||]_+,
\end{equation}
where $\mathcal{A}^{-}$ is the set of negative samples generated by randomly substituting one of the two pre-aligned entities. $||\cdot||$ denotes the $L_2$ vector norm. $[\cdot]_+=\max(0,\cdot)$ The distance of negative samples is expected to be larger than a margin $\lambda$, i.e., $|| \mathbf{h}_{i'} - \mathbf{h}_{j'}||>\lambda$. $\alpha_1$ is a hyper-parameter for balance.
	
Previous work usually uses the hidden outputs at the last layer as the final representations of entities, i.e. $\mathbf{h}_i=\mathbf{h}^{(L)}_i$ where $L$ denotes the number of layers. However, as discussed in Section \ref{sect:preli}, the representations of each layer all contribute to propagating alignment information. Therefore, we use the hidden representations of all layers. Formally, we have
\begin{equation}\label{eq:concat}
\mathbf{h}_i = \mathop{\oplus}\limits_{l=1}^L\text{norm}(\mathbf{h}_i^{(l)}),
\end{equation}
where $\oplus$ represents concatenation and $\text{norm}(\cdot)$ is the $L_2$ normalization for reducing the trivial optimization procedure of artificially increasing vector norm \cite{TransE}.

\subsection{Relation Semantics Modeling}
As KGs provide semantic relations between entities, it is natural to incorporate the semantics of the relational facts into entity modeling. As discussed in Section \ref{sect:preli}, R-GCN needs the structures of two KGs to be highly similar or relation alignment for entity alignment. Here, we borrow the translational assumption from TransE \cite{TransE}. To avoid overhead of parameters, we do not introduce additional relation-specific embeddings. The representation for $r$, denoted as $\mathbf{r}$, can be retrieved via its related entity embeddings:
\begin{equation} 
	\label{eq:relation}
	\mathbf{r} =\frac{1}{|\mathcal{T}_r|} \sum_{(s,o) \in \mathcal{T}_r}(\mathbf{h}_s-\mathbf{h}_o),
\end{equation}
where $\mathcal{T}_r$ is the subject-object entity pairs of relation $r$. Then we minimize the following relation loss for refinement:
\begin{equation} 
	\label{eq:trans_loss}
	\mathcal{L}_2 = \sum_{r\in \mathcal{R}}\frac{1}{|\mathcal{T}_r|} \sum_{(s,o) \in \mathcal{T}_r}||\mathbf{h}_s-\mathbf{h}_o-\mathbf{r}||,
\end{equation}
where $\mathcal{R}$ is the set of the total relations in the two KGs.

\subsection{Implementation}
Next, we introduce implementation details of AliNet.
\subsubsection{Objective}
The final objective of AliNet is the combination of the contrastive alignment loss and relation loss, aiming at injecting relation semantics to the preserved graph structures:
\begin{equation} 
	\label{eq:loss}
	\mathcal{L} = \mathcal{L}_1 + \alpha_2 \, \mathcal{L}_2,
\end{equation}
where $\alpha_2$ is a hyper-parameter to weight the two losses. The objective is optimized using the Adam optimizer. All the learnable parameters including the input feature vectors of entities are initialized by the Xavier initialization \cite{Xavier}. The adjacency information is a sparse matrix obtained from the relational triples $\mathcal{T}_1$ and $\mathcal{T}_2$. The neighborhood aggregation can be done by the sparse matrix multiplication between the adjacency matrix and the entity representation matrix, making the storage complexity linear to the number of entities and triples. 

\subsubsection{Generalization to $k$-hop neighborhood} 
Here we consider aggregating the neighborhood information within $k$ hops. Let $\rho_1(\mathbf{h}_{i,1}^{(l)},\mathbf{h}_{i,2}^{(l)})$ be the gating combination for the one-hop and two-hop neighborhood aggregation in Eq. (\ref{eq:gating}). We use $k-1$ gating functions to combine the information recursively:
\begin{equation} 
\label{eq:gating_k}
\mathbf{h}_{i}^{(l)} = \rho_{k-1}(\cdots\rho_2(\rho_1(\mathbf{h}_{i,1}^{(l)},\mathbf{h}_{i,2}^{(l)}),\mathbf{h}_{i,3}^{(l)})\cdots).
\end{equation}

\subsubsection{Neighborhood augmentation} 
The proposed gated multi-hop neighborhood aggregation expands the direct neighbors of an entity in an end-to-end manner. To further implement this idea, we propose a heuristic method to add edges among pre-aligned entities. Specifically, if two entities $i$ and $j$ of KG$_1$ have an edge while their counterparts $i'$ and $j'$ in KG$_2$ do not, we add an edge linking $i'$ and $j'$. The goal is to mitigate the non-isomorphism by adding such balanced edges.

\subsubsection{Alignment prediction} Once trained AliNet, we can predict entity alignment based on the nearest neighbor search among entity representations in the cross-KG scope. Given a source entity $i$ to be aligned in KG$_1$, its counterpart in KG$_2$ is: $i'=\arg \min_{j \in \mathcal{E}_2} \pi (\mathbf{h}_i, \mathbf{h}_{j}),$
where $\pi()$ is a distance measure such as Euclidean distance. Here we still use the combined representations to measure the distance of entity embeddings.

\section{Experiments}
In this section, we evaluate AliNet on the entity alignment task. The source code of AliNet is accessible online\footnote{\url{https://github.com/nju-websoft/AliNet}}.

\begin{table*}[!t]
	\centering
	\resizebox{\textwidth}{!}{
		\begin{tabular}{lcclcclcclcclccl}
				\toprule
				\multirow{2}{*}{Methods} & \multicolumn{3}{c}{DBP\textsubscript{ZH-EN}} & \multicolumn{3}{c}{DBP\textsubscript{JA-EN}} & \multicolumn{3}{c}{DBP\textsubscript{FR-EN}} & \multicolumn{3}{c}{DBP-WD} & \multicolumn{3}{c}{DBP-YG} \\
				\cmidrule(lr){2-4} \cmidrule(lr){5-7} \cmidrule(lr){8-10} \cmidrule(lr){11-13}  \cmidrule(lr){14-16} 
				& H@1 & H@10 & MRR & H@1 & H@10 & MRR & H@1 & H@10 & MRR & H@1 & H@10 & MRR & H@1 & H@10 & MRR  \\ \midrule
				MTransE \cite{MTransE} & 0.308 & 0.614 & 0.364 & 0.279 & 0.575 & 0.349 & 0.244 & 0.556 & 0.335 & 0.281 & 0.520 & 0.363 & 0.252 & 0.493 & 0.334 \\
				 IPTransE \cite{IPTransE} & 0.406 & 0.735 & 0.516 & 0.367 & 0.693 & 0.474 & 0.333 & 0.685 & 0.451 & 0.349 & 0.638 & 0.447 & 0.297 & 0.558 & 0.386 \\
				JAPE \cite{JAPE} & 0.412 & 0.745 & 0.490 & 0.363 & 0.685 & 0.476 & 0.324 & 0.667& 0.430 & 0.318 & 0.589 & 0.411 & 0.236 & 0.484 & 0.320 \\  
				AlignE \cite{BootEA} & 0.472 & 0.792 & 0.581 & 0.448  & 0.789 & 0.563 & 0.481 & 0.824 & 0.599 & 0.566 & 0.827 & 0.655  & 0.633 & 0.848 & 0.707 \\
				GCN-Align \cite{GCN_Align} & 0.413& 0.744 & 0.549 & 0.399 & 0.745 & 0.546 & 0.373 & 0.745 & 0.532 & 0.506 & 0.772 & 0.600 & 0.597 & 0.838 & 0.682 \\
				SEA \cite{SEA}& 0.424 & 0.796 & 0.548 & 0.385 & 0.783 & 0.518 & 0.400 & 0.797 & 0.533 & 0.518 & 0.802 & 0.616 & 0.516 & 0.736 & 0.592\\
				RSN \cite{RSN}& 0.508 & 0.745 & 0.591 & 0.507 & 0.737 & 0.590 & 0.516 & 0.768 & 0.605 & 0.607 & 0.793 & 0.673 & 0.689 & 0.878 & 0.756\\
				MuGCN \cite{MuGNN} & 0.494 & \textbf{0.844} & 0.611 & 0.501 & \textbf{0.857} & 0.621 & 0.495 & \textbf{0.870} & 0.621 & 0.616 & 0.897 & 0.714 & 0.741 & 0.937 & 0.810\\
				\midrule
				TransH \cite{TransH}& 0.377 & 0.711 & 0.490 & 0.339 & 0.681 & 0.462 & 0.313 & 0.668 & 0.433 & 0.351 & 0.641 & 0.450 & 0.314 & 0.574 & 0.402\\
				ConvE \cite{ConvE}& 0.169 & 0.329 & 0.224 & 0.192 & 0.343 & 0.246 & 0.240 & 0.459 & 0.316 & 0.403 & 0.628 & 0.483 & 0.503 & 0.736 & 0.582\\
				RotatE \cite{RotatE} & 0.485 & 0.788 & 0.589 & 0.442 & 0.761 & 0.550 & 0.345 & 0.738 & 0.476 & 0.479 & 0.776 & 0.579 & 0.599 & 0.835 & 0.680\\
				\midrule
				GCN \cite{GCN} & 0.487 & 0.790 & 0.559 & 0.507 & 0.805 & 0.618 & 0.508 & 0.808 & 0.628 & 0.613 & 0.850 & 0.698 & 0.733 & 0.909 & 0.796\\
				GAT \cite{GAT} & 0.418 & 0.667 & 0.508 & 0.446 & 0.695 & 0.537 & 0.442 & 0.731 & 0.546 & 0.540 & 0.781 & 0.625 & 0.563 & 0.806 & 0.648 \\
				R-GCN \cite{R-GCN} & 0.463 & 0.734 & 0.564 & 0.471 & 0.754 & 0.571 & 0.469 & 0.758 & 0.570 & 0.574 & 0.791 & 0.651 & 0.617 & 0.829 & 0.692\\
				\midrule
				AliNet (w/o rel. loss \& augment) & 0.511 & 0.798 & 0.611 & 0.527 & 0.794 & 0.622 & 0.520 & 0.848 & 0.635 & 0.642 & 0.877 & 0.726 & 0.745 & 0.918 & 0.806\\
				AliNet (w/o rel. loss) & 0.525 & 0.790 & 0.619 & 0.539 & 0.796 & 0.638 & 0.535 & 0.839 & 0.645 & 0.679 & 0.887 & 0.750 & 0.773 & 0.935 & 0.832\\
								AliNet & \textbf{0.539} & 0.826 & \textbf{0.628} & \textbf{0.549} & 0.831 & \textbf{0.645} & \textbf{0.552} & 0.852 & \textbf{0.657} & \textbf{0.690} & \textbf{0.908} & \textbf{0.766} & \textbf{0.786} & \textbf{0.943} & \textbf{0.841}\\
				\bottomrule
			\end{tabular}}
			\caption{Result comparison on entity alignment}
	\label{tab:ent_alignment}
\end{table*}

\subsection{Datasets}
Following the latest progress \cite{BootEA,MuGNN}, we use the following datasets and training-test splits.
\begin{itemize} 
	\item DBP15K \cite{JAPE} has three datasets built from multi-lingual DBpedia, namely DBP\textsubscript{ZH-EN} (Chinese-English), DBP\textsubscript{JA-EN} (Japanese-English) and DBP\textsubscript{FR-EN} (French-English). Each dataset has $15,000$ reference entity alignment and about four hundred thousand triples.
	\item DWY100K \cite{BootEA} are extracted from DBpedia, Wikidata and YAGO3. It has two datasets, namely DBP-WD (DBpedia-Wikidata) and DBP-YG (DBpedia-YAGO3). Each dataset has $100,000$ reference entity alignment and more than nine hundred thousand triples.
\end{itemize}  

\subsection{Comparative Models}
We compare with recent embedding-based entity alignment models: MTransE \cite{MTransE}, IPTransE \cite{IPTransE}, JAPE \cite{JAPE}, AlignE \cite{BootEA}, GCN-Align \cite{GCN_Align}, SEA \cite{SEA}, RSN \cite{RSN} and MuGNN \cite{MuGNN}. Note that some recent GNN-based models like GMNN \cite{KGMatching} and RDGCN \cite{RDGCN} incorporate the surface information of entities into their representations. As our model solely relies on structure information, we do not take these models into comparison. For ablation study, we develop three variants of AliNet, i.e., AliNet (w/o rel. loss) that does not optimize the relation loss, AliNet (w/o rel. loss \& augment.) that does not employ the relation loss and neighborhood augmentation, and the full model AliNet.

For comprehensive comparison, we also choose some KG embedding models and GNN variants as baselines. Conventional KG embedding models are usually evaluated on the task of link prediction. However, as studied in \cite{MultiKE}, some of them can also be used for entity alignment. We select TransH \cite{TransH}, ConvE \cite{ConvE} and RotatE \cite{RotatE}. TransE \cite{TransE} has already been exploited for entity alignment by MTransE and IPTransE. For GNNs, we choose GCN \cite{GCN}, GAT \cite{GAT} and R-CGN \cite{R-GCN} as baselines. The re-tuned versions of GCN, GAT and R-GCN are implemented by ourselves following the same pipeline as AliNet for fair comparison. 

\subsection{Implementation Details}
We search among the following values for hyper-parameters, i.e., the learning rate in $\{0.0001,0.0005,0.001,0.005,0.01\}$, $\alpha_1$ in $\{0.1,0.2,\dots,0.5\}$, $\alpha_2$ in $\{0.01,0.05,0.1,0.2\}$, $\lambda$ in $\{1.0,1.1,\dots,2.0\}$, the hidden representation dimension of each layer in $\{100,200,300,400,500\}$, the number of layers $L$ in $\{1,2,3,4\}$, and the number of negative alignment pairs in $\{5,10,15,20\}$. The selected setting is that $\lambda=1.5$, $\alpha_1=0.1$, $\alpha_2=0.01$. The learning rate is $0.001$. The batch size for DBP15K is $4,500$, and for DWY100K is $10,000$. We stack two AliNet layers ($L=2$) and each layer combines the one-hop and two-hop information ($k=2$). The dimensions of three layers (including the input layer) are $500$, $400$ and $300$, respectively. The activation function for neighborhood aggregation is $\text{tanh}()$, and the one for the gating mechanism is $\text{ReLU}()$. We sample 10 negative samples for each pre-aligned entity pair. We use early stopping to terminate training based on the Hits@1 performance with a patience of 5 epochs. We use CSLS \cite{Word_Translation} for nearest neighbor search.

Following convention, we report the Hits@1, Hits@10~and MRR (mean reciprocal rank) results to assess entity alignment performance. Higher Hits@$1$, Hits@$10$ and MRR scores indicate better performance. Note that, the Hits@1 is equivalent to precision. As the nearest neighbor search can always find a counterpart for each entity to be aligned, the recall and F1-measure also have the same value as Hits@1. 

\subsection{Main Results}
We present the entity alignment results in Table \ref{tab:ent_alignment}. We can see that AliNet outperforms the state-of-the-art structure-based embedding models for entity alignment by Hits@1 and MRR. For example, on DBP\textsubscript{FR-EN}, AliNet achieves a gain of $0.036$ by Hits@1 compared with RSN, and $0.057$ against MuGNN. We think that these results have demonstrated the superiority of AliNet. As the DBP15K datasets are extracted from the multi-lingual DBpedia, the schema heterogeneity of them is much weaker than that of DWY100K which are extracted from different KGs. AliNet also achieves the best Hits@10 results on DWY100K, demonstrating its practicability. We find that the neighborhood augmentation method leads AliNet a gain of $0.012-0.037$ by Hits@1. This is because it can reduce the non-isomorphism in the neighborhood structures of pre-aligned entities. The results further support our motivation of mitigating the non-isomorphism in KG structures for entity alignment. AliNet shows better performance than AliNet (w/o rel. loss), showing the effect of the relation loss.
 
In comparison to the re-tuned GNN variants GCN, GAT and R-GCN, AliNet also achieves better performance. As the GCN baseline has the same training process as AliNet and the difference only lies in the choice of GNN layers (i.e., GCN layers for one-hop neighborhood aggregation or AliNet layers for both one-hop and two-hop neighborhood aggregation), these results can demonstrate the effectiveness of integrating multi-hop information. Both GAT and R-GCN fail to outperform GCN. We attribute such observation to that the direct neighbors of an entity are less dissimilar than distant neighbors, and hence may not require an attention-based neighborhood aggregation to select relevant neighboring entities. This is also the reason for choosing GCN layers rather than GAT layers in the one-hop neighborhood aggregation of AliNet. For R-GCN, as discussed in Section \ref{sect:preli}, it cannot well capture the similarity of neighborhood structures of counterpart entities.  We also observe that the GCN baseline outperforms many other embedding-based entity alignment models including another GCN variant GCN-Align. This again validates the effectiveness of our framework. 

\subsection{Analyses}
\subsubsection{Aggregation strategies of multi-hop neighborhood}
The underlying idea of AliNet is to extend the neighborhood of entities by attentively aggregating multi-hop neighborhood with the gating mechanism. To gain a deep insight into this point, we further design three variants of AliNet using different strategies to aggregate multi-hop neighborhood. The first one, denoted as AliNet (mix), borrows the idea from MixHop \cite{MixHop}, which has a similar motivation for node classification in general graphs. It takes the two-hop neighbors as one-hop and uses GCN layers to directly aggregate such mixed neighborhood information. The second one, denoted as AliNet (add), replaces the gating mechanism with addition operator. In the last variant AliNet (gat), we replace the proposed attention mechanism with GAT \cite{GAT}. Due to space limitation, we only show the results on DBP15K in Table \ref{tab:multi_hop_results}. We find that AliNet (mix) fails to achieve promising performance, which indicates that using GCN layers for two-hop neighborhood aggregation is not effective because it would introduce much noise information. AliNet (add) does not show very satisfactory results because addition cannot selectively combine important representations across dimensions like the gating mechanism. AliNet (gat) also achieves slightly lower performance than AliNet, showing the effect of our attention mechanism. Based on these results and the performance of the GCN baseline shown in Table \ref{tab:ent_alignment}, we can come to the conclusion that the multi-hop neighborhood information indeed contributes to entity alignment while the gating and attention mechanisms are crucial to capture important information in distant neighbors.

\begin{table}[!t]
	\centering
		\resizebox{\linewidth}{!}{
		\begin{tabular}{lcclcclccl}
				\toprule
				\multirow{2}{*}{Methods} & \multicolumn{3}{c}{DBP\textsubscript{ZH-EN}} & \multicolumn{3}{c}{DBP\textsubscript{JA-EN}} & \multicolumn{3}{c}{DBP\textsubscript{FR-EN}}\\
				\cmidrule(lr){2-4} \cmidrule(lr){5-7} \cmidrule(lr){8-10} 
				& H@1 & H@10 & MRR & H@1 & H@10 & MRR & H@1 & H@10 & MRR \\ \midrule
				AliNet (mix) & 0.227 & 0.611 & 0.350 & 0.294 & 0.696 & 0.426 & 0.258 & 0.674 & 0.391 \\
				AliNet (add) & 0.498 & 0.801 & 0.602 & 0.515 & 0.813 & 0.618 & 0.501 & 0.839 & 0.585 \\
				AliNet (gat) & 0.517 & 0.803 & 0.618 & 0.531 & 0.810 & 0.632 & 0.523 & 0.845 & 0.636\\
				AliNet & \textbf{0.539} & \textbf{0.826} & \textbf{0.628} & \textbf{0.549} & \textbf{0.831} & \textbf{0.645} & \textbf{0.552} & \textbf{0.852} & \textbf{0.657} \\
				\bottomrule
			\end{tabular}}
	\caption{Results on DBP15K w.r.t. aggregation strategies}
	\label{tab:multi_hop_results}
\end{table}

\begin{figure}[!t]
	\center
	\includegraphics[width=0.99\linewidth]{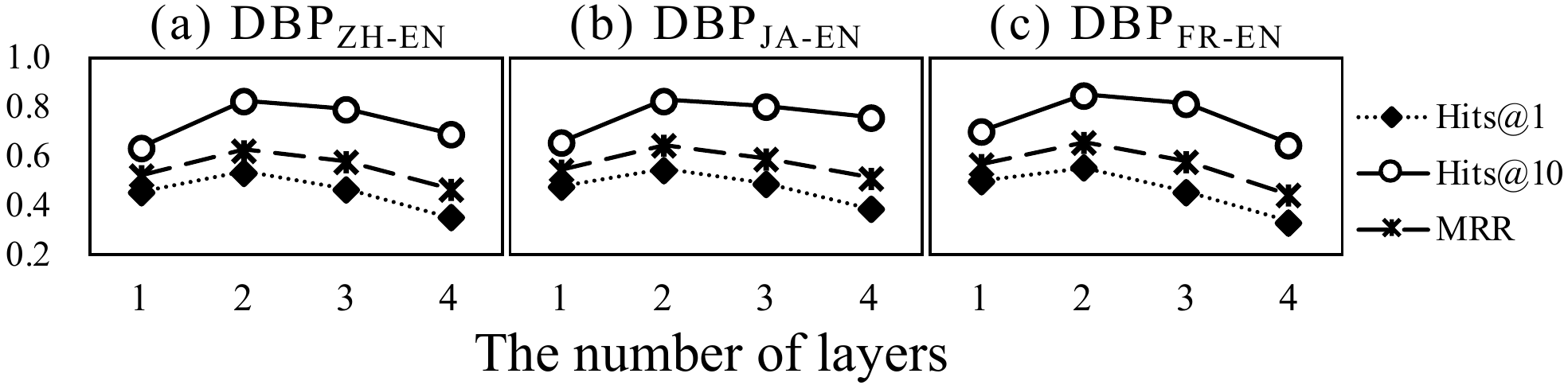}
	\caption{Results on DBP15K w.r.t. the number of layers.}
	\label{fig:layer_num}
\end{figure}

\subsubsection{Impact of the number of layers and choice of $k$}
We first report the results of AliNet with 1 to 4 layers on DBP15K in Figure \ref{fig:layer_num}. AliNet with 2 layers achieves the best performance over all the three metrics. We observe that when AliNet has more layers, its performance declines as well. Although more layers allow AliNet to indirectly capture more distant neighborhood information by layer-to-layer propagation, such distant neighbors would introduce much noise and lead to more non-isomorphic neighborhood structures. Besides, we further show the results of the two-layer AliNet that considers different hops of neighborhood information in each layer in Table \ref{tab:k_hop_results}. We can see that considering two-hop neighborhood leads to the best results. This is similarly attributed to the aforementioned reasons regarding aggregation of multi-hop neighbors. This is further verified by an analysis about DBP15K. For example, in DBP\textsubscript{ZH-EN}, each Chinese entity has 6.6 one-hop neighbors on average and this number for each English entity is 8.6. However, between their one-hop neighbors, there are only 4.5 pairs of counterpart entities, leaving 2.1 Chinese one-hop neighbors and 4.1 English ones unaligned. If considering two-hop neighbors, the numbers of unaligned one-hop neighbors are reduced to 0.5 for Chinese and 0.9 for English, respectively. The numbers have less room to be reduced by introducing more distant neighbors. This suggests us that aggregating two-hop neighborhood information is enough.

\begin{table}[!t]
	\centering
	\resizebox{\linewidth}{!}{
		\begin{tabular}{lcclcclccl}
				\toprule
				\multirow{2}{*}{Methods} & \multicolumn{3}{c}{DBP\textsubscript{ZH-EN}} & \multicolumn{3}{c}{DBP\textsubscript{JA-EN}} & \multicolumn{3}{c}{DBP\textsubscript{FR-EN}}\\
				\cmidrule(lr){2-4} \cmidrule(lr){5-7} \cmidrule(lr){8-10} 
				& H@1 & H@10 & MRR & H@1 & H@10 & MRR & H@1 & H@10 & MRR \\ \midrule
				GCN & 0.487 & 0.790 & 0.559 & 0.507 & 0.805 & 0.618 & 0.508 & 0.808 & 0.628 \\
				AliNet & \textbf{0.539} & \textbf{0.826} & \textbf{0.628} & \textbf{0.549} & \textbf{0.831} & \textbf{0.645} & \textbf{0.552} & \textbf{0.852} & \textbf{0.657} \\
				AliNet ($k=3$) & 0.461 & 0.786 & 0.571 & 0.484 & 0.802 & 0.590 & 0.450 & 0.813 & 0.575  \\
				AliNet ($k=4$) & 0.386 & 0.721 & 0.501 & 0.407 & 0.706 & 0.516 & 0.373 & 0.745 & 0.499\\
				\bottomrule
			\end{tabular}}
	\caption{Results on DBP15K w.r.t. $k$ values}
	\label{tab:k_hop_results}
\end{table}

\begin{figure}[!t]
	\center
	\includegraphics[width=0.9999\linewidth]{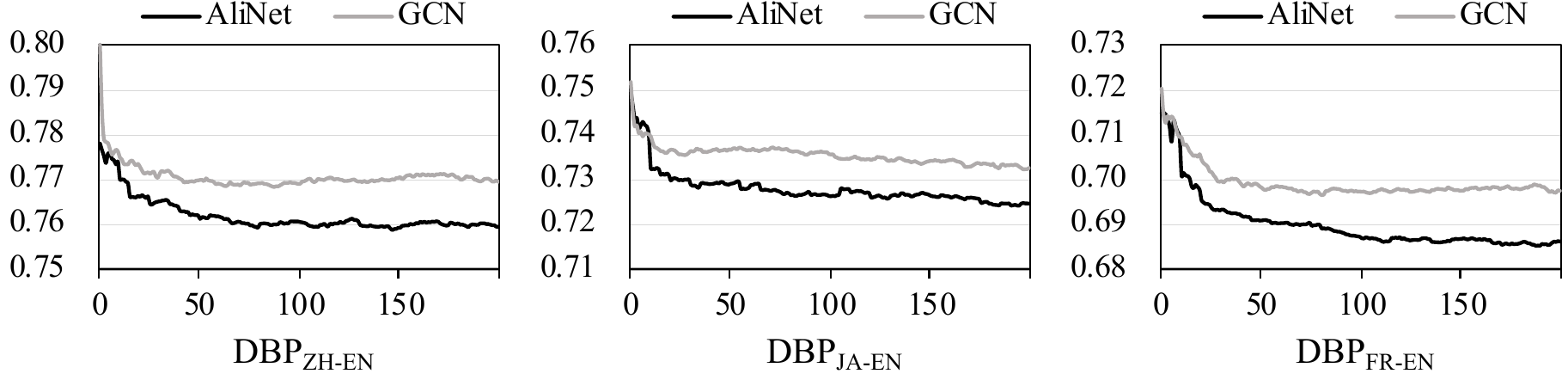}
	\caption{Average OC of one-hop neighbor sets of correct alignment during the first 200 training epochs on DBP15K.}
	\label{fig:overlap}
\end{figure}

\subsubsection{Analysis of neighborhood overlap} Furthermore, we make an empirical statistics on the overlap coefficient (OC) of the one-hop neighbors for each pair of counterpart entities in the correctly-found alignment. A~high OC value for two entities means that they have a large overlap between their one-hop neighbors. We predict entity alignment and compute the average OC values of the correctly-predicted alignment every epoch. Figure \ref{fig:overlap} shows the value changes during the first 200 training epochs of AliNet and GCN. We find that the values display a decreasing trend. This indicates that it is relatively easy for GNN-based models to find the counterpart entities having a large proportion of common one-hop neighbors. The OC values of AliNet are smaller than those of GCN. This indicates that AliNet can effectively align the entities with smaller overlap in their one-hop neighbors. 

\subsubsection{Performance based on different layers} 
\begin{figure}[!t]
	\center
	\includegraphics[width=0.9999\linewidth]{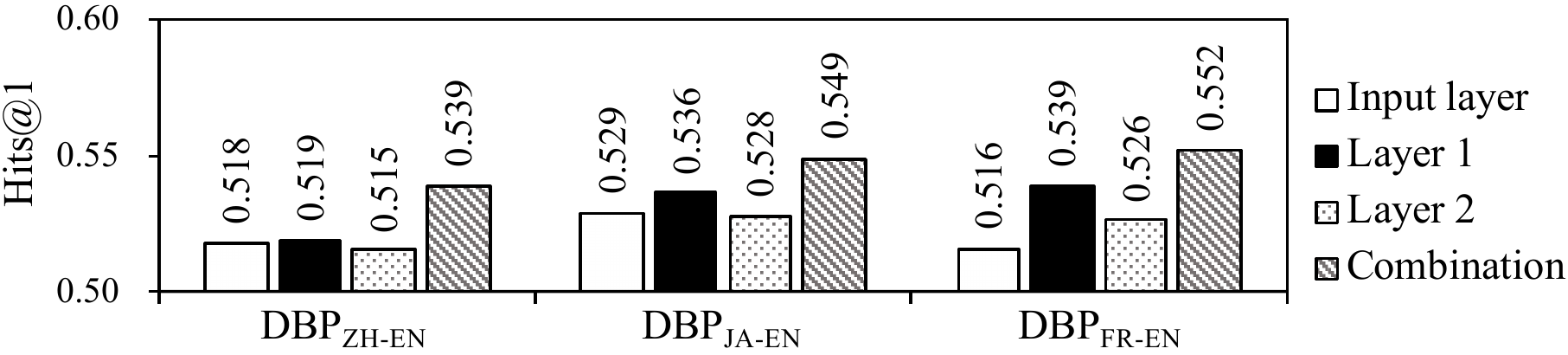}
	\caption{Hits@1 results w.r.t. different layers on DBP15K.}
	\label{fig:layer}
\end{figure}

In AliNet, we propose to use the combined representations of all layers as the final entity representations for predicting entity alignment. Here, we further examine the performance based on layer-specific entity representations. We report the entity alignment results on DBP15K in Figure \ref{fig:layer} due to space limitation. The input layer is the randomly initialized feature vectors for entities to be tuned in AliNet. On top of this, we stack two AliNet layers (i.e., Layer 1 and Layer 2). ``Combination'' means the combined representations computed by Eq. (\ref{eq:concat}). We can see that the representations of different layers show different performance of entity alignment. Layer 1 shows the best results among the three layers. As expected, the combined representations finally outperform the layer-specific representations. 

\section{Related Work}
We hereby discuss related work to this paper. Particularly, as we have covered the majority of GNN-based entity alignment models in Section \ref{sect:preli}, we focus on discussing other families of models here. Most other models, such as MTransE \cite{MTransE}, IPTransE \cite{IPTransE}, JAPE \cite{JAPE}, AlignE and BootEA \cite{BootEA}, NAEA \cite{NAEA} as well as OTEA \cite{OTEA}, use TransE \cite{TransE} to learn entity embeddings. Meanwhile they learn a linear mapping or minimize the distance between the embeddings of pre-aligned entities. On top of KG structures, some work like KDCoE \cite{KDCoE}, AttrE \cite{AttrE} and MultiKE \cite{MultiKE}, incorporates additional profile information of entities such as textual descriptions and literal names for KG embedding. Differently, AliNet exploits the basic graph structures without using additional information. To further improve entity alignment performance, IPTransE, BootEA, KDCoE and NAEA use semi-supervised learning.

We also notice the recent work \cite{GMN} that uses GNNs for comparing the similarity of two graphs. Differently, we focus on the node-level rather than graph-level similarity comparison. \cite{KampffmeyerCLWZ19} captures the hierarchical structures of entities by introducing hypernym-hyponym links between nodes and their ancestors/descendants. Our work is also related to KG embedding that aims at learning vector representations for KG completion. There are translational models such as TransE \cite{TransE}, TransH \cite{TransH} and TransR \cite{TransR}, bilinear models such as ComplEx \cite{ComplEx}, SimplE \cite{SimplE} and RotatE \cite{RotatE}, and deep models such as ConvE \cite{ConvE}, R-GCN \cite{R-GCN} and RSN \cite{RSN}. We refer interested readers to the recent survey \cite{KRL} for more details on KG embedding.

\section{Conclusion and Future Work}
In this paper, we propose AliNet for entity alignment, aiming at mitigating the non-isomorphism among the neighborhood structures of counterpart entities in an end-to-end manner. AliNet captures the neighborhood information within multiple hops by a gating mechanism in each layer. It employs an attention mechanism for multi-hop neighborhood aggregation to reduce noises. We further propose a relation loss to enhance the expressiveness of AliNet. Our experiments on five datasets demonstrate the effectiveness of AliNet. For future work, we plan to incorporate side information of entities in other modalities into the preserved graph structures.

\subsubsection*{Acknowledgments}
This work is supported by the National Key R\&D Program of China (No. 2018YFB1004300), the National Natural Science Foundation of China (No. 61872172), and the Key R\&D Program of Jiangsu Science and Technology Department (No. BE2018131).

\bibliographystyle{aaai}
\bibliography{aaai_reference}

\end{document}